\documentclass{article}

\usepackage{PRIMEarxiv}

\usepackage[utf8]{inputenc} 
\usepackage[T1]{fontenc}    
\usepackage{hyperref}       
\usepackage{url}            
\usepackage{booktabs}       
\usepackage{amsfonts}       
\usepackage{nicefrac}       
\usepackage{microtype}      
\usepackage{lipsum}
\usepackage{fancyhdr}       
\usepackage{graphicx}       
\graphicspath{{media/}}     

\usepackage{amsmath}
\usepackage{bm}
\usepackage{mathtools}
\usepackage{amssymb}
\usepackage{longtable}
\usepackage{comment}
\usepackage{tikz}
\usepackage{xcolor}

\newtheorem{definition}{Definition}
\newtheorem{proposition}{Proposition}

\title{\Large Designing Control Barrier Function via Probabilistic Enumeration for Safe Reinforcement Learning Navigation}

\author{
  Luca Marzari$^{1,*}$, Francesco Trotti$^{2,}$\thanks{Marzari and Trotti contributed equally to the paper.}\;, Enrico Marchesini$^{3}$ and Alessandro Farinelli$^{1}$
  \\
  $1$ Department of Computer Science, University of Verona, Verona, Italy.\\
  $2$ Department of Engineering for Innovation Medicine, University of Verona, Verona, Italy.\\
  $3$ Massachusetts Institute of Technology, Boston, USA.\\
  Contact authors: \textit{luca.marzari@univr.it}, \textit{francesco.trotti@univr.it}.}

\begin{document}
\maketitle

\begin{abstract}
Achieving safe autonomous navigation systems is critical for deploying robots in dynamic and uncertain real-world environments. In this paper, we propose a hierarchical control framework leveraging neural network verification techniques to design control barrier functions (CBFs) and policy correction mechanisms that ensure safe reinforcement learning navigation policies. 
Our approach relies on probabilistic enumeration to identify unsafe regions of operation, which are then used to construct a safe CBF-based control layer applicable to arbitrary policies. We validate our framework both in simulation and on a real robot, using a standard mobile robot benchmark and a highly dynamic aquatic environmental monitoring task. These experiments demonstrate the ability of the proposed solution to correct unsafe actions while preserving efficient navigation behavior. Our results show the promise of developing hierarchical verification-based systems to enable safe and robust navigation behaviors in complex scenarios.
\end{abstract}


\section{Introduction}

Achieving safe autonomous navigation is a crucial requirement to deploy robots in the real world, where human interactions and expensive hardware are often involved. For example, robots can assist humans in monitoring and mitigating pollution in a variety of environmental sustainability scenarios (e.g., aquatic ecosystems \cite{aquaculture}) or perform complex autonomous navigation in precision agriculture tasks \cite{pierce1999aspects}. Traditionally, these applications are based on manual human intervention, which is costly and struggles to capture real-time changes reliably. These are challenges that a drone equipped with a safe autonomous navigation stack can successfully address \cite{aquaculture2}.
 To this end, online model-based learning methods embedding hierarchical control architectures have been used to deal with uncertainty in various navigation tasks \cite{best2019dec, trotti2024markov, trotti2024path, trotti2023online}. Nevertheless, these methods rely on extensive domain knowledge, which could be difficult to obtain in practice. To address this issue, deep reinforcement learning (DRL) algorithms have been employed to learn how to navigate in complex and unknown (i.e., mapless) environments \cite{ji2023safety, marzari2023online, omnisafe, marzari2024improving}. However, DRL-based policies are modeled as deep neural networks (DNNs), which are known to be vulnerable to adversarial inputs---small perturbations to the input state leading to unexpected and unsafe actions \cite{adversarial, TACAS}. Two main lines of work have been investigated to address such an issue. On the one hand, formal verification (FV) of neural networks can certify safety, providing rigorous theoretical guarantees that a DNN-based navigation policy lies within predefined safety constraints \cite{LiuSurvey}. However, FV scales poorly and novel probabilistic verification approaches have recently gained traction to address the limits of formal methods \cite{proven,marzari2023dnn}. In detail, probabilistic enumeration \cite{marzari2024enumerating} can identify regions of the state space (i.e., parts of the environment) where a DNN exhibits unsafe behavior with a high degree of confidence, but this information has not been exploited to correct unsafe navigation behaviors. 
On the other hand, control barrier functions (CBFs) \cite{cbf} also provide a principled way to enforce safety constraints by specifying a function that characterizes a safe set, ensuring its forward invariance \cite{cbf}. Hence, CBFs can potentially guarantee that navigation trajectories remain within pre-defined safety regions. However, designing effective CBFs for high-dimensional, complex environments remains challenging when dealing with uncertainty and complex dynamics. 
To mitigate these issues, recent work has explored the use of neural CBFs and their verification \cite{liu2023safe, so2024train, CBFverification}. Despite the different lines of research, ensuring the successful deployment of safe policies and correcting their decisions when detecting an unsafe situation remains an open research question.

\begin{figure}[t]
    \centering
    \includegraphics[width=0.7\linewidth]{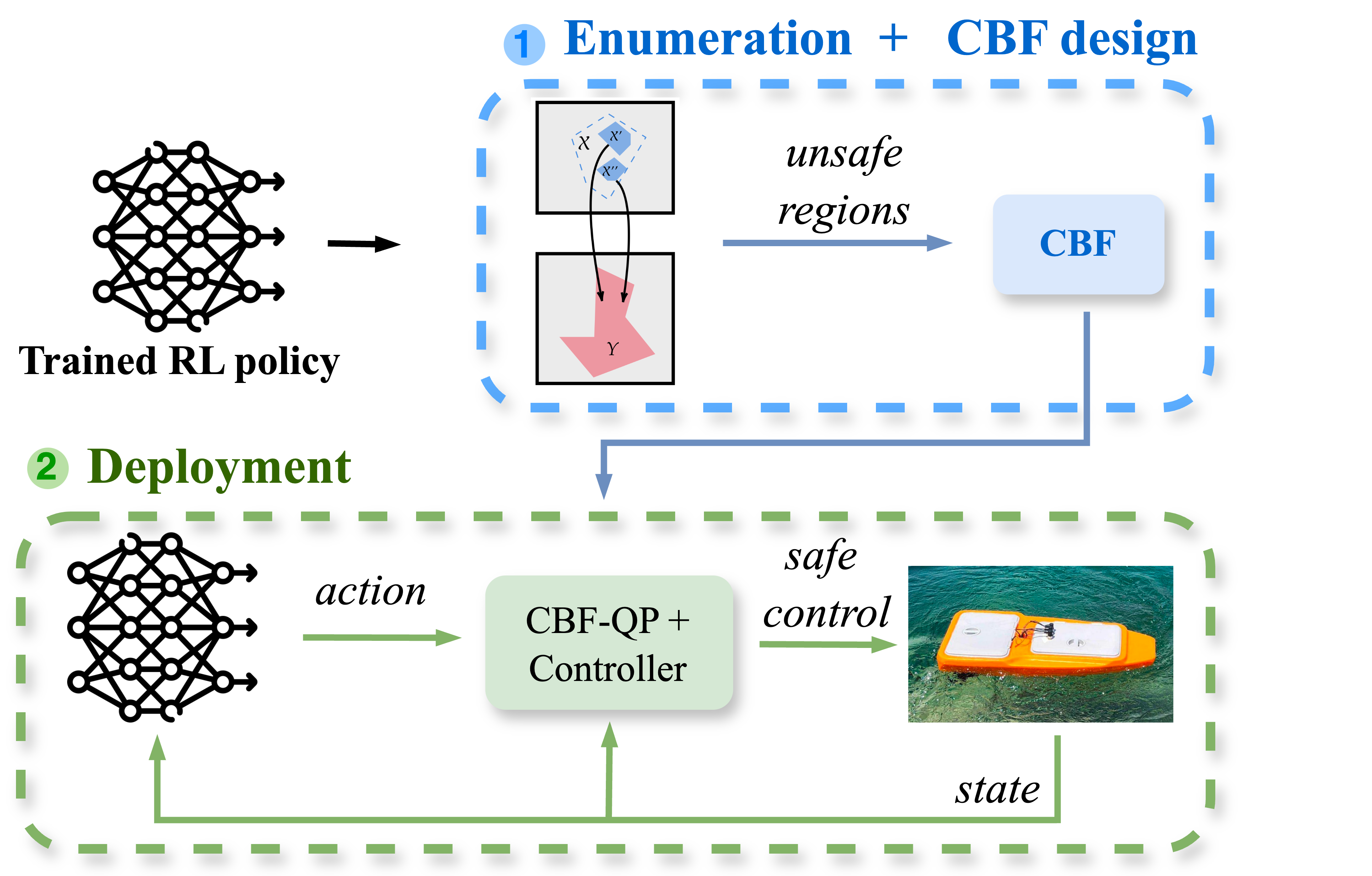}
    \caption{Overview of the proposed approach.}
    \label{fig:overview}
\end{figure}

We address this gap by leveraging the advantages of probabilistic verification and CBF-based control, introducing a framework that guarantees safe DRL-based navigation policies and correct policy actions in potentially unsafe situations (Fig. \ref{fig:overview}). 
At a high level, we compute a safe navigation set by first identifying unsafe regions through offline probabilistic enumeration \cite{marzari2024enumerating} and then removing them from the state space of the DRL policy. We then design a CBF-based control mechanism to ensure the agent avoids both these pre-identified unsafe regions and any obstacles detected during navigation---ensuring the agent remains within the precomputed safe set. At deployment time, a quadratic programming (QP) optimization incorporating the CBF's safety constraints evaluates the policy's action, determining whether the chosen action keeps the agent within the precomputed safe set or not. If the action violates any safety constraints, a low-level controller modifies the action to ensure the agent returns to the safe set while maintaining effective navigation behaviors.

A key advantage of our approach is that the enumeration component, which employ an abstract interval representation to identify the (un)safe regions, is agnostic to the specific navigation environment where the agent is deployed, thus enhancing generalization. Furthermore, the proposed framework works on top of arbitrary DRL policies and it is orthogonal to both the training and verification strategies employed by a neural CBF for the execution of safe policies.

To confirm the effectiveness of the proposed approach, we first validated our strategy in two different Unity-based simulated navigation scenarios using a mobile Turtlebot3 and an aquatic drone for water monitoring. We then performed a real-world deployment of a trained DRL policy to validate the correctness of our pipeline. Our empirical results demonstrate that the proposed method enables safe and scalable autonomous robotic navigation, achieving zero violations of safety restrictions at deployment while increasing navigation effectiveness.
\section{Background and Preliminaries}
In this section, we cover the main foundations related to the research topics considered by our framework.

\subsection{Reinforcement Learning and Constrained Approaches}
Navigation tasks are typically modeled as a Markov decision process (MDP), represented by a tuple $<\mathcal{S},\mathcal{A}, R,\mathcal{P},\gamma>$ where $\mathcal{S} \in \mathbb{R}^d$ is the state space, $\mathcal{A}\in \mathbb{R}^2$ is the action space---which in this work consists of linear and angular velocities---$\mathcal{P}: \mathcal{S} \times \mathcal{A} \to \mathcal{S}$ is the
state transition function and $R: \mathcal{S} \times \mathcal{A} \to \mathbb{R}$ is a reward function. Given a stationary policy $\pi \in \prod$, an agent aims to maximize $J_{r}^{\pi} = \mathbb{E}_{\tau \sim \pi} \left[\sum_{t=0}^\infty \gamma^t R(s_t, a_t)\right]$, i.e., the expected discounted return for each trajectory $\tau = (s_0, a_0, \cdots, s_n,a_n)$, where $\gamma \in (0, 1]$ is a discount factor used to regulate the impact of future rewards. To consider safety, MDPs are typically extended to constrained MDPs (CMDPs) and incorporate a set of constraints during the training using one or more cost functions $C: \mathcal{S} \times \mathcal{A} \to [0, 1]$. Hence, constrained DRL algorithms aim at maximizing the expected return $J_r^{\pi}$ while maintaining costs under hard-coded thresholds $\mathbf{t}$:
$\max_{\pi \in \Pi} J_{r}^{\pi} \quad\text{s.t.}\quad J_{C}^{\pi} \leq \mathbf{t}$, with $J_{C}^{\pi} = \mathbb{E}_{\tau \sim \pi} [\sum_{t=0}^\infty \gamma^t C(s_t, a_t)]$. However, CMDPs present non-negligible limitations, such as the necessity of a parametrization of the policy (i.e., they are not applicable to value-based methods) \cite{Lagrangian}. To address this issue, recent work  \cite{marchesini2023navigation} showed how incorporating penalties into the reward effectively influences policy toward safer behaviors while avoiding the limitations introduced by constrained approaches. Nonetheless, these methods only guarantee safety in expectation, making them unsuitable to ensure safe deployment.

\subsection{Formal Verification of Neural Networks}\label{prelim:FV}

\begin{figure}[h!]
    \centering
    \includegraphics[width=0.7\linewidth]{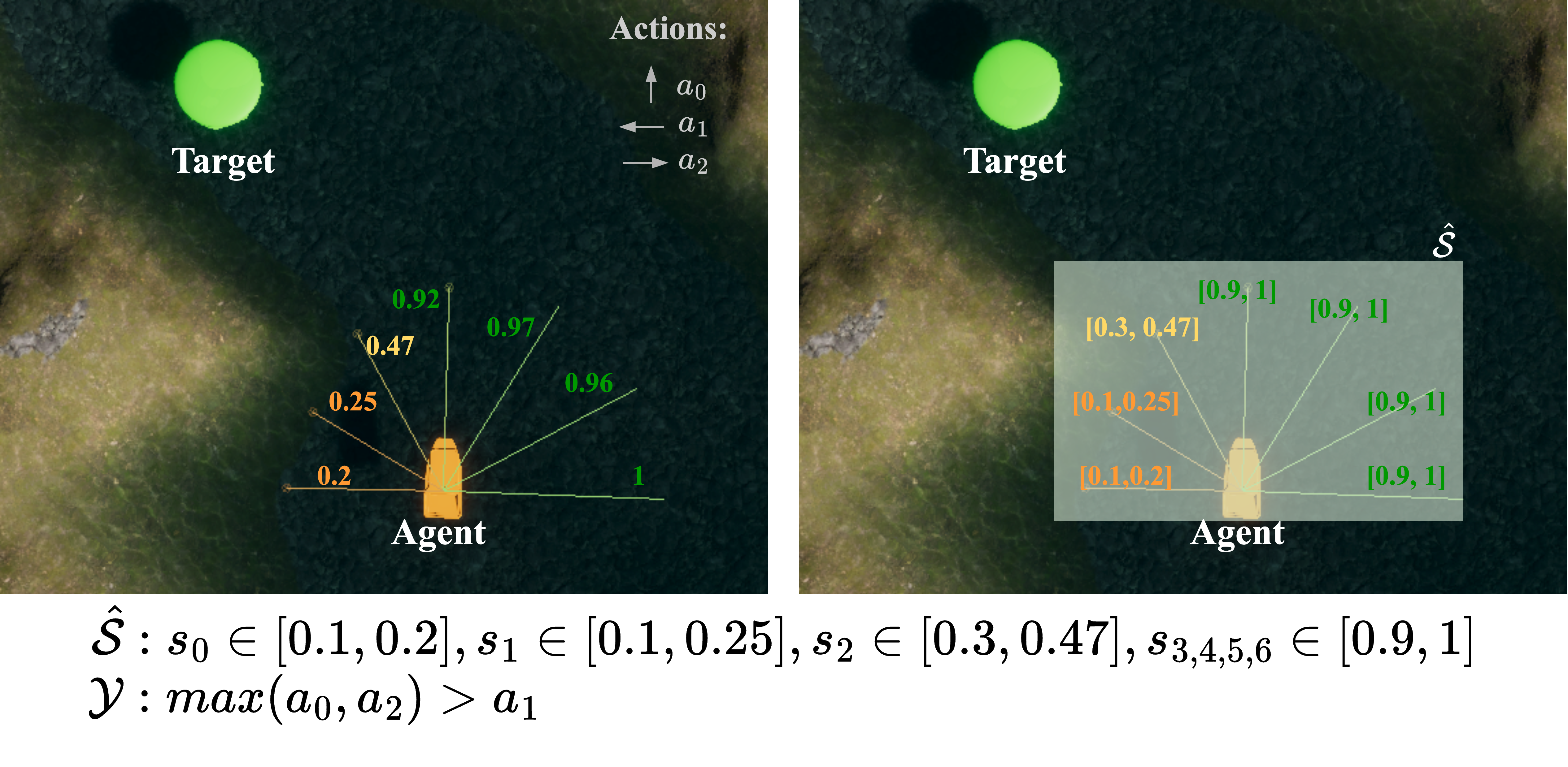}
    \vspace{-3mm}
    \caption{Illustrative example of a safety property in aquatic mapless navigation: the left shows an unsafe state near the coastline, while the right defines a region of the state space $\hat{\mathcal{S}} \subset \mathcal{S}$ to verify the agents never choose a leftward movement.}
    \label{fig:safety_props}
    \vspace{-5mm}
\end{figure}

Given a trained policy represented by a neural network $\mathcal{N}: \mathcal{S} \to \mathcal{A}$, formal verification checks a safety property defined in terms of input-output relationships for $\mathcal{N}$ \cite{LiuSurvey}. The general idea is to select a subset of the state space $\hat{\mathcal{S}} \subset \mathcal{S}$, define $\mathcal{Y}$ that specifies the desired (safe) output action for that portion of the state space, and verify it. Fig. \ref{fig:safety_props} shows an explanatory example of a safety property for the aquatic navigation task. 

In detail, a FV tool propagates the intervals defined by $\hat{\mathcal{S}}$ through $\mathcal{N}$, performing a layer-by-layer reachability analysis to compute the output reachable set $\mathcal{R}(\hat{\mathcal{S}})$, i.e., the set of all possible outputs selected by the DNN for the region $\hat{\mathcal{S}}$. The tool then verifies whether $\mathcal{R}(\hat{\mathcal{S}}) \subseteq \mathcal{Y}$---indicating that the policy satisfies the safety property for all states in $\hat{\mathcal{S}}$. If even a single state $\bm{s}=[s_0,\ldots,s_n]^T \in \hat{\mathcal{S}}$ produces an output $a$ that falls outside $\mathcal{Y}$, the property is considered violated.  
However, due to the nonlinear nature of deep neural networks, determining whether such a violation exists is an NP-complete problem. Additionally, the binary nature of FV results (i.e., whether a violation occurs or not) does not provide sufficient safety information. For example, one model might exhibit only a small unsafe region around a specific counterexample, while another might have multiple widespread violations. To address the scalability and expressiveness of FV, recent works \cite{marzari2024enumerating} propose to probabilistically enumerate all the portions of $\hat{\mathcal{S}}$ respecting or violating the safety desiderata, thus quantifying the rate at which agents either satisfy or do not satisfy the DNN input-output relationships. We emphasize that, although the enumeration strategy used in our work is inherently probabilistic, its application to identifying unsafe regions reflects a conservative approach—misclassifying a portion of a safe area as unsafe is acceptable, whereas the opposite would compromise safety. In Sec. \ref{subsubsec: forward_inv} and \ref{subsubsec: cbf}, we will show how to exploit the enumeration results to design the CBF.

\subsection{Control Barrier Function}

Consider an affine nonlinear system of the form:  
\begin{equation}
\label{eq: affine_sys}
    \dot{x} = f(x) + g(x)u,
\end{equation}  
where $ x \in \mathcal{X} \subseteq \mathbb{R}^{n} $ represents the state variable within the dynamics model state space $\mathcal{X}$, and $ u \in \mathcal{U} \subseteq \mathbb{R}^{m} $ is the control input within the control input space $\mathcal{U}$. The functions $ f(x) $ and $ g(x) $ are nonlinear mappings from $ \mathbb{R}^{n} $ to $ \mathbb{R}^{n} $ and are assumed to be Lipschitz continuous. In this setting, a CBF is defined as follows.

\begin{definition}[Control Barrier Function \cite{cbf}]
\label{def: cbf}
    Let $\mathcal{C} \subset \mathbb{R}^{n}$ be the set defined by a continuously differentiable function $h: \mathbb{R}^{n} \rightarrow \mathbb{R}$ such that
    \begin{equation}
        \begin{split}
            \mathcal{C} &= \{x \in \mathbb{R}^{n}: h(x) \ge 0 \}, \\
            \partial\mathcal{C} &= \{x \in \mathbb{R}^{n}: h(x) = 0 \}, \\
            \text{Int}(\mathcal{C}) &= \{x \in \mathbb{R}^{n}: h(x) > 0 \}. \\
        \end{split}
    \end{equation}
We say that $h$ is a control barrier function if 
$\nabla h(x) \ne 0$ for all $x \in \partial\mathcal{C}$ and there exists a class-$\mathcal{K}$ function $\alpha$, such that for all $x \in \mathcal{C}, \exists u $ such that $L_{f}h(x) + L_{g}h(x)u \ge -\alpha(h(x)),$
with $L_{f}h(x):= \nabla h(x)^{T}f(x)$ and $L_{g}h(x):= \nabla h(x)^{T}g(x)$. 
\end{definition}

This formalization will be used in Sec \ref{subsubsec: cbf} to define the CBF for navigation policies.  
 
\subsection{Dynamic Models}
\label{subsec: dyn}

The design of the proposed pipeline requires the definition of the dynamic models to compute a corrective action for the navigation policy. Specifically, in this work, we consider (i) the kinematic equations of a mobile robot (Turtlebot3) and (ii) the nonlinear dynamics model of an aquatic drone. 
\subsubsection{Mobile robot}
\label{subsubsec: dyn_MB}
The kinematics equations of the robot are represented by $\dot{p}_x = v_1 \cos{\theta}, \quad \dot{p}_y = v_1 \sin{\theta}, \quad   \dot{\theta} = \omega_3$, where $p_x, p_y, \theta$ are the positions and orientation of the robot and $\bm{u} = [v_1, \omega_3]$ are the controlled linear and angular velocities.
 
\subsubsection{Aquatic drone}
\label{subsubsec: dyn_AD}
The aquatic drone motion is described following the 6-degrees-of-freedom nonlinear dynamic model and the fluid dynamics coefficients are computed through spline interpolation using a high-fidelity simulator following common formalisms in the literature \cite{trotti2024towards, gierusz2016modelling}. Hence, the aquatic drone state and control vector are defined as:
\begin{equation} \label{eq:dyn_state}
    \small
    \begin{split}
    \bm{x} &= [v_1 \quad v_2 \quad v_3 \quad \omega_1 \quad \omega_2 \quad \omega_3 \quad p_{x} \quad p_{y} \quad p_{z} \quad \phi \quad  \theta \quad \psi]^{T},\\
    \bm{u} &= [\delta_{l} \quad \delta_{r}],
    \end{split}
\end{equation}
where $v_1, v_2, v_3$ are linear velocities and $\omega_1, \omega_2, \omega_3$ are the rate of the angles in the body frame, respectively. With $p_{x}, p_{y}, p_{z}$ and $\phi, \theta, \psi$, we refer to the boat position and the Euler angles in the Earth frame. Regarding the control variables, $\delta_l$ and $\delta_r$ represent the left and right thrusts of the boat.
For readability purposes, in the next we call $a=(\cos{\psi}\sin{\theta}\sin{\phi} - \sin{\psi}\cos{\phi}), b=(\cos{\psi}\sin{\theta}\cos{\phi} + \sin{\psi}\sin{\phi}), c=(\sin{\psi}\sin{\theta}\sin{\phi} + \cos{\psi}\cos{\phi}), d=(\sin{\psi}\sin{\theta}\cos{\phi} - \cos{\psi}\sin{\phi}), e = \cos{\theta}\sin{\phi}, f = \cos{\theta}\cos{\phi}$. Hence, following the affine dynamic system of the form \eqref{eq: affine_sys}, the kinematic components of the positions and orientation (Euler angles) in Earth frame of the dynamic model can be expressed in the following form:

\begin{equation}
\label{eq: fx_gx_kin}
\begin{aligned}
    \begin{bmatrix}
        \dot{p}_{x} \\
        \dot{p}_{y} \\
        \dot{p}_{z} \\
        \dot{\phi} \\
        \dot{\theta} \\
        \dot{\psi} \\
    \end{bmatrix} 
    &=
    \underbrace{
    \begin{bmatrix}
        v_2 a + v_3 b \\
        v_2 c + v_2 d \\
        v_2 e + v_3 f \\
        \omega_{1} + \omega_{2}\tan{\theta}\sin{\phi} \\
        \omega_{2}\cos{\phi} \\
        \frac{\omega_{2}\sin{\phi}}{\cos{\theta}}
    \end{bmatrix}}_{f(x)}
    +
    \underbrace{
    \begin{bmatrix}
        \cos{\psi}\cos{\theta} & 0 \\
        \sin{\psi}\cos{\phi} & 0 \\
        -\sin{\theta} & 0 \\
        0 & \tan{\theta}\cos{\phi} \\
        0 & -\sin{\phi} \\
        0 & \frac{\cos{\phi}}{\cos{\theta}}
    \end{bmatrix}}_{g(x)}
    r.
\end{aligned}
\end{equation}

Concerning the dynamic components of the linear and angular accelerations in the body frame, we have:

\begin{equation}
\label{eq: fx_gx_dyn}
\begin{aligned}
    \begin{bmatrix}
        \dot{v}_{1} \\
        \dot{v}_{2} \\
        \dot{v}_{3} \\
        \dot{\omega}_{1} \\
        \dot{\omega}_{2} \\
        \dot{\omega}_{3} \\
    \end{bmatrix} 
    &=
    \underbrace{
    \begin{bmatrix}
       - \omega_{2}v_3 + \omega_{3}v_2 - g  \sin(\theta) \\
        - \omega_{3}v_1 + \omega_{1}v_3 + g  \sin(\phi)  \cos(\theta) \\
        - \omega_{1}v_2 + \omega_{2}v_1 + g  \cos(\phi)  \cos(\theta) \\
        (c_1\omega_{2} +c_2\omega_{1})\omega_{2} \\
        c_5\omega_{1}\omega_{3} - c_6(\omega_{1}^2 - \omega_{3}^2) \\
        (c_8\omega_{1}-c_2\omega_{3})\omega_{2}
    \end{bmatrix}}_{f(x)}
    +
    \underbrace{
    \begin{bmatrix}
       X/m & 0 \\
        Y/m & 0 \\
       Z/m & 0 \\
        0 & K \\
        0 & M \\
        0 & N
    \end{bmatrix}}_{g(x)}
    u,
\end{aligned}
\end{equation}
with $X, Y, Z, K, M, N$ defined as:
\begin{equation}
\begin{tabular}{l c c}
$X = (\delta_r + \delta_l) - F_{x}$ & $\, \quad \,$ & $F_x =\frac{1}{2} \rho v_{1}^2 C_{Fx} A_x$,\\  
$Y = - F_{y}$ & $\, \quad \,$ & $F_y =\frac{1}{2} \rho v_{2}^2 C_{Fy} A_y$, \\
$Z = F_{b} - F_{z}$ & $\, \quad \,$ & $F_z = \frac{1}{2} \rho v_{3}^2 C_{Fz} A_z$,  \\
$K = - M_{x}- C_k(\phi)  F_{b}$ & $\, \quad \,$ & $M_x = \frac{1}{2} \rho \omega_{1}^2 C_{Mx} I_x$,\\  
$M = - M_{y} - C_m(\theta)  F_{b}$ & $\, \quad \,$ & $M_y = \frac{1}{2} \rho \omega_{2}^2 C_{My} I_y$, \\
$N = \frac{1}{2}(\delta_r - \delta_l)  d - M_{z}$ & $\, \quad \,$ & $M_z = \frac{1}{2} \rho \omega_{3}^2 C_{Mz} I_z$.\\
\end{tabular}
\end{equation}
In detail, $F_{b} = \rho g V_{\text{sub}}(p_z)C_b$ with $V_{\text{sub}}$ representing the submerged volume, $C_{(\cdot)}$ are the drag coefficient on the translational and moment on the tree axes and the bouncing drag respectively, while $A_x, A_y, A_z$ are the contact areas with the water, and $d$ is the distance between the two motors.

\section{Problem Statement}

\begin{figure}[h!]
    \centering
    \includegraphics[width=0.8\linewidth]{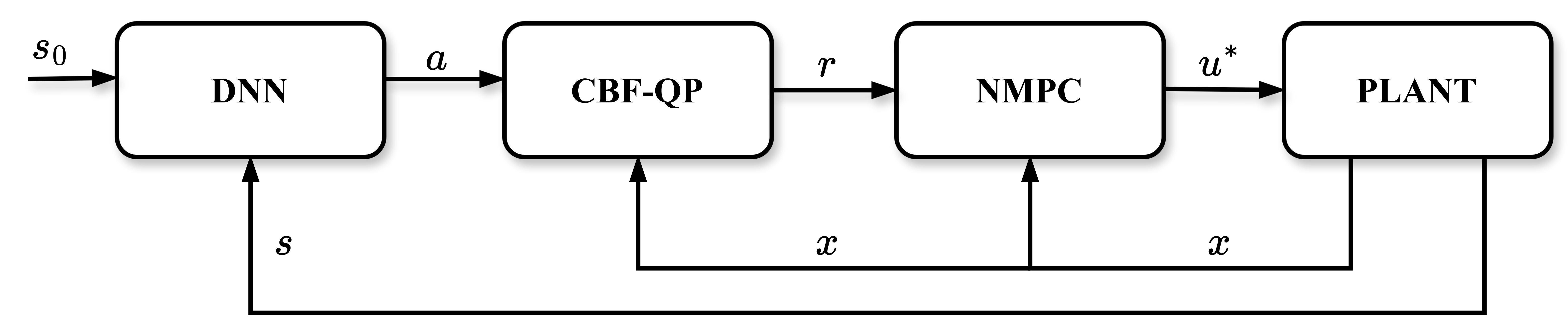}
    \caption{Block diagram of the hierarchical architecture.}
    \label{fig: complete}
\end{figure}

To ensure safe and successful deployment of the DRL-based navigation policy, we define a CBF to provide a reference for a low-level controller that corrects the policy action (if necessary) to satisfy the safety constraints. To this end, we first investigate how it is possible to synthesize a forward invariant safe set $\mathcal{C}$ for a DRL agent that has to navigate within a given environment. The overall process is summarized in Fig. \ref{fig: complete}.

In detail, we consider a neural network $\mathcal{N}$ that encodes a DRL policy $\pi$ and define an input state $s$ as part of the dynamics state vector $\bm{x}$ (Eq. \ref{eq:dyn_state}). The input space of the policy consists of: (i) $15$ sparse beam scans for the boat or LiDAR for the mobile robot, each representing the mean value measurements sampled within a cone in the range $[0, 360]$ degree; (ii) the agent’s position $p = [p_{x}, p_{y}, p_{z}]^{T}$ and orientation $\eta = [\phi, \theta, \psi]^T$ (i.e., odometry values); and (iii) the target's relative position and orientation (i.e., distance and heading).  Given a state $s$, the policy $\pi$ predicts an action $a = [v_1^{\text{dnn}}, \omega_3^{\text{dnn}}]^T$, representing the agent's linear and angular reference velocities. To ensure that this action respects safety criteria, we aim to keep the agent within safe regions of the state space (in the context of navigation, these safety criteria relate to avoiding collisions with obstacles). Our intuition is to identify these regions and design a safe set $\mathcal{C}$ by exploiting the enumeration process. Given the computed safe set, we then formalize a CBF $h(x)$ to enforce safety constraints in the agent’s trajectory toward the target. Our approach involves correcting the policy actions by solving a QP problem that incorporates the CBF constraints. This procedure generates reference velocities $r$, which serve as inputs for an optimal low-level controller based on nonlinear model predictive control (NMPC) \cite{NMPC}. The latter computes the optimal control input $u^*$ for the plant, which evolves using the dynamic \eqref{eq: fx_gx_dyn} providing both the new state $\bm{x}$ for the controller and CBF, and $s$ for the policy.

\section{Methodology}
\subsection{Compute the Forward Invariant Safe Set} \label{subsubsec: forward_inv}

To construct the forward-invariant safe set, we identify the regions where the agent does not adhere to safety specifications within a subset of the state space of interest, denoted as $\hat{\mathcal{S}}$. To this end, we rely on the probabilistic enumeration tool proposed by \cite{marzari2024enumerating}. In detail, this process begins at training time, where we leverage an indicator cost function to detect collisions. This function allows us to identify portions (or regions, interchangeably) of the state space where the agent is prone to unsafe behaviors. In particular, at training time, we collect the state, i.e., the agent's range of inputs (e.g., lidars, odometry, etc), which translates into unsafe actions \cite{marzari2023online}. After training, we manually define a safety property for each (state, action) pair by analyzing the neighborhood of the corresponding unsafe state, denoted as $\hat{\mathcal{S}}_i$ (for $i = \{1, \ldots, n\}$, with $n$ is the total number of unsafe (state, action) pairs). Within this area, we probably enumerate the subset of unsafe regions, represented as $\overline{\mathcal{C}}_i$. Importantly, by encoding the neighborhoods of unsafe (state, action) pairs, the approach becomes independent of the specific environment used to train the agent, enabling the identification and mapping of unsafe behaviors to different, potentially unseen scenarios.

\begin{figure}[t]
    \centering
    \includegraphics[width=0.8\linewidth]{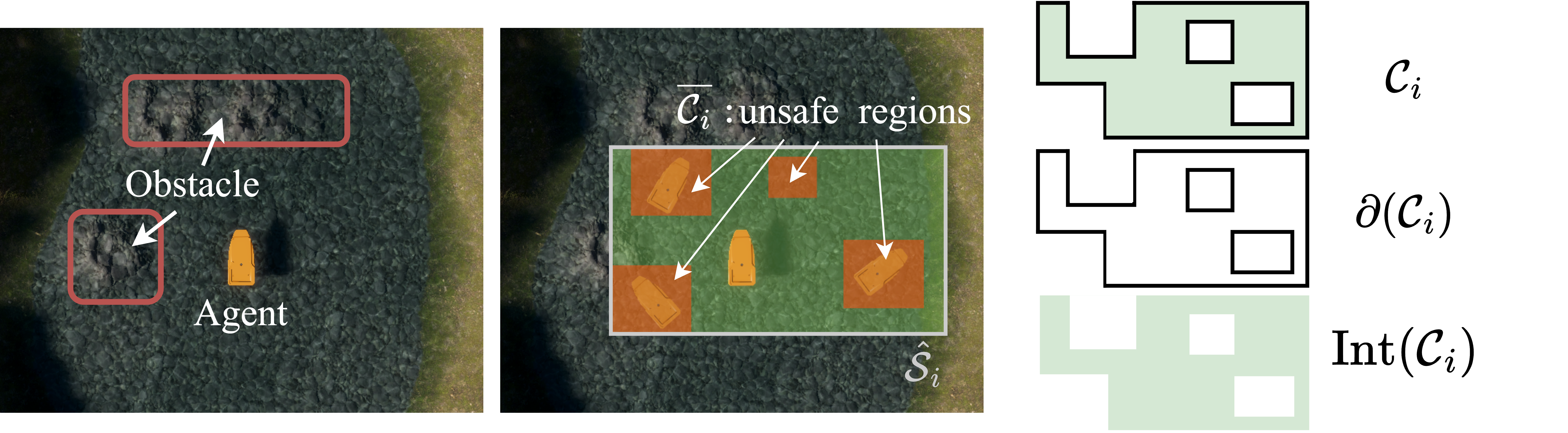}
    \caption{Explanatory example of the relationship between the enumeration process and the CBF. On the left, we report an unsafe situation where the agent has two obstacles on the left and the coastline on the right. In the center, we define the input region to be verified in green, and we enumerate the unsafe regions where the agent has a high probability of colliding. Finally, on the right, we report the translation of the enumeration result into the CBF's formulation.}
    \label{fig:example_enum}
\end{figure}

Fig. \ref{fig:example_enum} shows an explanatory example with the results of the verification process for a region of interest $\hat{\mathcal{S}_i}$, where we identify the set of safe regions as $\mathcal{C}_i = \hat{\mathcal{S}_i} \setminus \overline{\mathcal{C}_i}.$
As illustrated in the right figure, each safe subset $\mathcal{C}_i$ is derived by removing a verified unsafe region $\overline{\mathcal{C}_i}$, which consists of a finite union of closed sets from the compact state space subset $\hat{\mathcal{S}_i}$. Notably, removing a finite union of closed sets from a compact set preserves compactness, thus making each $\mathcal{C}_i$ a valid safe subset for a CBF \cite{cbf,lin1996smooth}. Hence, by taking the union of all such safe subsets, we obtain a globally valid safe set $\mathcal{C}$ for the CBF. Formally, 

\begin{proposition}\label{prop:valid_safe_set}
    The set $\mathcal{C} = \bigcup_i \mathcal{C}_i$ is a valid safe set for a control barrier function.
\end{proposition}

In the next section, we show that from a safe set $\mathcal{C}$ as in Prop. \ref{prop:valid_safe_set} there exists a continuously differentiable function $h: \mathbb{R}^n \to \mathbb{R}$ such that $h(x) \geq 0$ for all $x \in \mathcal{C}$.

\subsection{Control Barrier Function and Low-Level Controller}
After the computation of the safe set $\mathcal{C}$, we need to define a CBF that keeps the agent in this set. To this end, starting from the dynamic models of the agent (see Sec. \ref{subsec: dyn}), we propose a strategy to compensate for the policy actions and ensure safety. Hence, we design a hierarchical approach that forces the agent to comply with the constraint generated by the verification process. 
 
\subsubsection{Control Barrier Function}
\label{subsubsec: cbf}
Given the linear and angular reference velocities produced by the policy, our goal is to rescale them to satisfy the safety constraints. To achieve this, we design a CBF that enforces the position constraint defined by $\mathcal{C}$. We then incorporate this CBF into a QP formulation that adjusts the policy’s reference values, ensuring the resulting actions keep the agent within the safe set.
Given the constraint on the position, we focus only on the kinematic components of the dynamic model defined in equations \eqref{eq: fx_gx_kin}. 
Therefore, we have the safe set computed by the enumeration $\mathcal{C} = \bigcup_i \mathcal{C}_i =\{ x \in \mathbb{R}^{n} : h(x) \ge 0\}$, 
with the control barrier function designed as follows:
\begin{equation}
\label{eq: cbf}
    h(x) = || p - p_{\text{obs}} ||^{2} - d_{\text{safe}}^{2},
\end{equation}
where $p = [p_{x}, p_{y}, p_{z}]^{T}$ is the position of the agent, $p_{\text{obs}} = [p_x^{\text{obs}}, p_y^{\text{obs}}, p_z^{\text{obs}}]^{T}$ is the position of the obstacle detected by the sensor and $d_{\text{safe}} = \max(\sigma, ||p - p_{\text{area}}||^2)$ 
is a safe distance to the unsafe areas. In particular, $p_{\text{area}}$ is the centroid of each enumerated unsafe area, and $\sigma$ is a safe threshold manually defined and based on the sensor precision. 
Hence, as stated in \textit{Def.} \ref{def: cbf}, the CBF constraint enforces forward invariance of the set $\mathcal{C}$, defining $\dot{h}(x) + \alpha h(x) \ge 0.$ By computing the time derivative of $h(x)$ as $ \dot{h}(x) = \frac{\partial h}{\partial x} \dot{x}$ we obtain
\begin{equation}
    \nabla h(x) f(x) + \nabla h(x) g(x) r + \alpha(h(x)) \ge 0,
\end{equation}
where $f(x)$ and $g(x)$ are the matrices in equation \eqref{eq: fx_gx_kin}, and:
\begin{equation}
    \small
     \nabla h(x) = \Big[2(p_x - p_x^{\text{obs}}),2(p_y - p_y^{\text{obs}}),2(p_z - p_z^{\text{obs}}),0,0,0\Big],
\end{equation}

At this point, we have the constraint on the linear and angular reference velocities, and we can formulate a QP problem to get a modulation action when the policy action does not comply with the constraint.
We model the reference variable $r$ as the sum of the policy DNN and CBF as follows:
\begin{equation}
    r = 
    \begin{bmatrix}
        v_{1}^{\text{dnn}} + v_{1}^{\text{cbf}}\\
        \omega_{3}^{\text{dnn}} + \omega_{3}^{\text{cbf}}\\
    \end{bmatrix}.
\end{equation}
To compute the contribution of the CBF on the linear and angular reference velocities ($r^{\text{cbf}} = [v_{1}^{\text{cbf}}, \omega_{3}^{\text{cbf}}]^{T}$), we solve an optimization problem formalized as QP scaling the given action. In detail, the QP formulation is:
\begin{align*}
\label{eq: opt_prob}
\displaystyle &r^{\text{cbf}}=\min_{v_{1}^{\text{cbf}}, \omega_{3}^{\text{cbf}}} \qquad ||v_{1}^{\text{cbf}}||^{2} + ||\omega_{3}^{\text{cbf}}||^{2}, \notag \\
& \text{subject to:} \notag \\
& \nabla h(x) f(x) + \nabla h(x) g(x)     
        \begin{bmatrix}
        v_{1}^{\text{dnn}} + v_{1}^{\text{cbf}}\\
        \omega_{3}^{\text{dnn}} + \omega_{3}^{\text{cbf}} \\
    \end{bmatrix} + \alpha(h(x)) \ge 0. \notag  \\
\end{align*}
In this way, if the action of the DNN complies with the constraint, the explicit solution to the QP problem does not provide a contribution on $v_{1}^{\text{cbf}}$ and $\omega_{3}^{\text{cbf}}$. Otherwise, the QP problem provides some linear and angular velocities in order to modulate the action of the DNN to comply with the constraints. 
Importantly, we note that the kinematic model of the Turtlebot3 mobile robot described in Sec. \ref{subsubsec: dyn_MB} and the barrier function $h(x)$ defined as \eqref{eq: cbf}, leads to the reformulated QP problem as:
\begin{align*}
\displaystyle &r^{\text{cbf}}=\min_{v^{\text{cbf}}, \omega^{\text{cbf}}} \qquad ||v^{\text{cbf}}||^{2} + ||\omega^{\text{cbf}}||^{2}, \notag \\
& \text{subject to:} \notag \\
&\begin{bmatrix}
        2(p_x - p_{x}^{\text{obs}})\\
        2(p_y - p_{y}^{\text{obs}}) \\
    \end{bmatrix}^{T}
    \begin{bmatrix}
        \cos\theta \\
        \sin\theta \\
    \end{bmatrix} 
    (v^{\text{dnn}} + v^{\text{cbf}}) + \\ & 
    \begin{bmatrix}
        2(p_x - p_{x}^{\text{obs}})\\
        2(p_y - p_{y}^{\text{obs}}) \\
    \end{bmatrix}^{T}
    \begin{bmatrix}
        -\sin\theta \\
        \cos\theta \\
    \end{bmatrix} (\omega^{\text{dnn}} + \omega^{\text{cbf}})
    + \alpha(h(x)) \ge 0. \notag
\end{align*}

\subsubsection{Low-level controller}\label{subsubsec: low_level}
After computing safe reference values, we focus on reaching them in an optimal way by designing an NMPC to manage the control variables $\delta_l, \delta_r$. 
To this end, we focus on the dynamic model defined by equation \eqref{eq: fx_gx_dyn}. We formalize the NMPC to compute the optimal control $u^{*}_k)$ over a finite time horizon $[k, k+H]$ at discrete time $t_k = k \Delta t,  k \in \mathbb{N}$ as

\begin{align}
u^{*}_{k} = & \displaystyle \min_{u_{k}, \dots, u_{k+H}} \qquad \sum_{i=0}^{H-1} e_{k+i|k}^{T}Qe_{k+i|k} + u_{k+i|k}^{T}Ru_{k+i|k}, \notag \\
& \text{subject to:} \notag \\
& x_{k+i+1|k} = f(x_{k+i|k}) + g(x_{k+i|k}) u_{k+i|k},  \notag \\
& u_{k+i|k}, \in\mathcal{U} \qquad u_{\min} \le u_{k+i|k} \le u_{\max} \notag \\
& \qquad \qquad \qquad \qquad \qquad \qquad \forall i \in \{ 0, \dots, H-1\}, \notag  \\
& x_{k|k} = x_{k}, \notag 
\end{align}
where $x_{k}$ are the initial conditions at time $k$ equal to the actual state of the plant $x_k$ and  $H$ is the horizon of NMPC. The error is deefined as $e_{k+i|k} = ||x_{k+i|k} - r||$ and the immediate cost function and $ Q \in \mathbb{R}^{n \times n} $ and $ R \in \mathbb{R}^{m \times m} $ are symmetric positive semi-definite matrices. Note that we indicate the predicted evolution of the state with $k+i|k$ where the right $k$ is the actual time, and the left $k+i$ is the time in the prediction horizon. 
\begin{figure}[h!]
  \centering
  \begin{minipage}[c]{0.45\textwidth}
    \centering
    \includegraphics[width=0.8\linewidth]{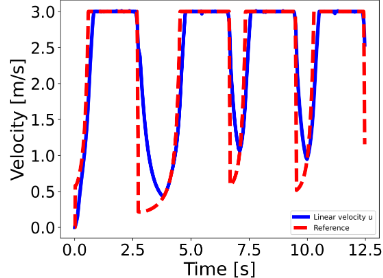} 
  \end{minipage}
  \hspace{0.3cm}
  \begin{minipage}[c]{0.45\textwidth}
    \centering
    \includegraphics[width=0.8\linewidth]{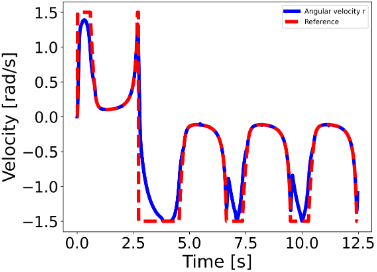} 
  \end{minipage}
    \caption{Linear and angular velocities tracking via NMPC. Red dashed lines are the reference $r$ while the blue lines are the optimal control action of the NMPC $u^{*}$.}
  \label{fig: nmpc}
\end{figure}

To confirm the correctness of the proposed solution, Fig. \ref{fig: nmpc} shows the action of the nonlinear model predictive control employed in our pipeline in order to track the reference (dashed red lines in the plot) provided by the policy modulated via QP problems. Specifically, the figure illustrates the control action on the linear and angular velocities of the boat (blue lines in the plot), indicating that the controller manages and minimizes the tracking error during the evolution. For readability purposes, we only report a short execution of the NMPC with different changing reference values.

\section{Empirical Evaluation}\label{sec:results}

\begin{figure}[h!]
  \centering
  \begin{minipage}[c]{0.24\textwidth}
    \centering
    \includegraphics[width=0.8\linewidth]{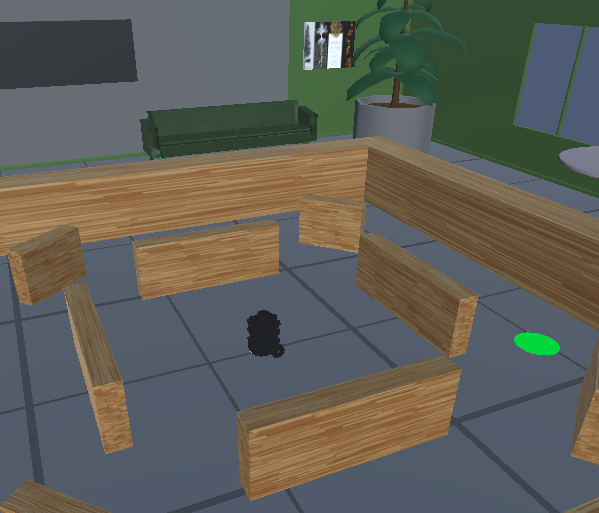} 
  \end{minipage}
  \hfill
  \begin{minipage}[c]{0.24\textwidth}
    \centering
    \includegraphics[width=0.8\linewidth]{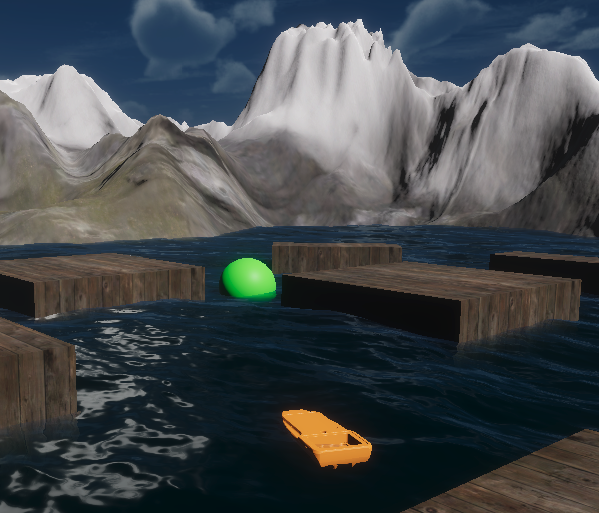} 
  \end{minipage}
    \hfill
  \begin{minipage}[c]{0.24\textwidth}
    \centering
    \includegraphics[width=0.8\linewidth]{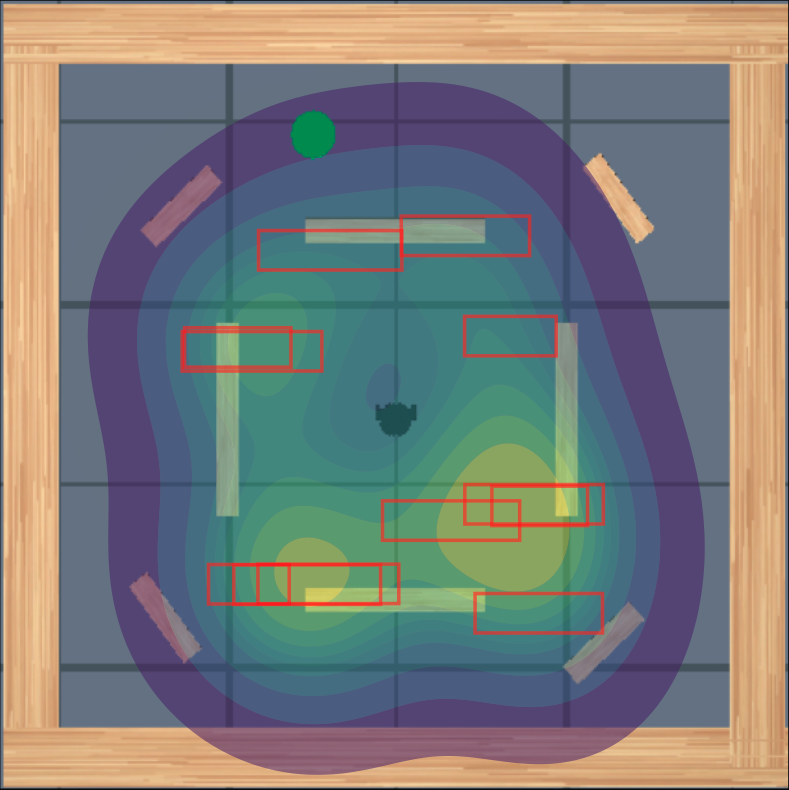} 
  \end{minipage}
    \hfill
  \begin{minipage}[c]{0.24\textwidth}
    \centering
    \includegraphics[width=0.8\linewidth]{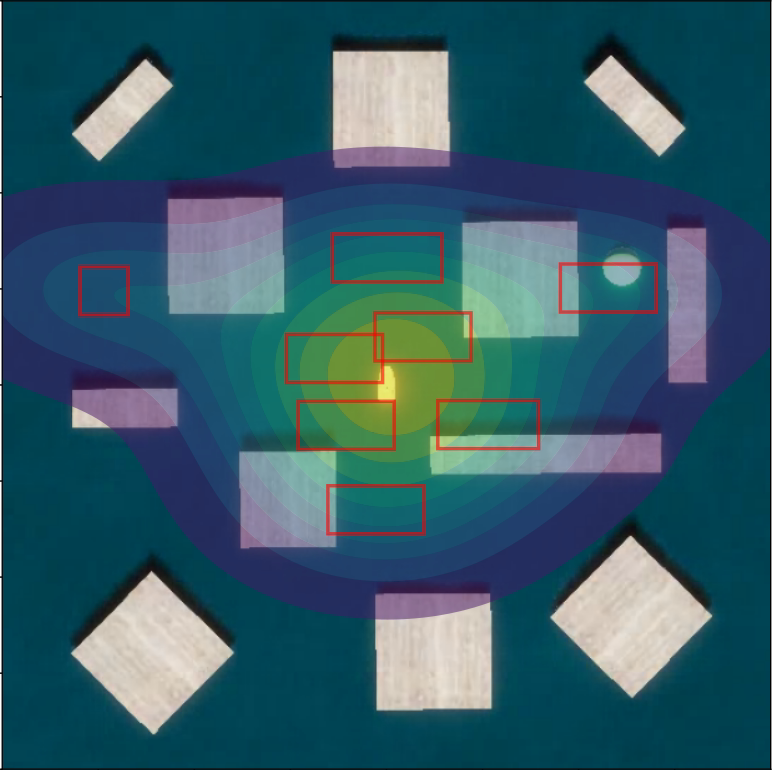} 
  \end{minipage}
   
  \caption{Indoor mobile navigation scenario and aquatic navigation task. On the right, density maps of unsafe regions identified post-training via the enumeration strategy.}
   \label{fig: envs_and_density}
\end{figure}

To validate the proposed autonomous safe navigation framework, we conducted a thorough evaluation in simulation and real scenarios. Our goal is to assess, given a DRL-trained policy for a mapless navigation task, whether: (i) our verification-guided CBF layer can consistently correct unsafe actions (i.e., ensure zero collisions) while preserving goal-reaching efficiency; and (ii) empirically investigate if such a layer can unstuck the agents when the policy ends up in local-minima. This is particularly relevant since a policy trained to satisfy stringent safety constraints typically prefers actions that keep the agent still to satisfy the strict thresholds imposed at training time. 
For the DRL policies, we consider popular unconstrained, penalty-based, and constrained algorithms that have been widely employed in safe navigation tasks \cite{ji2023safety, omnisafe, marchesini2023navigation} - PPO \cite{PPO}, PPO\_penalty and its Lagragian (safe) version, PPOLag \cite{Lagrangian}. In detail, PPO does not receive information regarding collisions, PPO\_penalty uses a $-0.01$ reward penalty to incentivize collision-free behaviors, and PPOLag uses a cost threshold set to 0 (aiming to achieve no collisions). We train these policies on an RTX 2070 and an i7-9700k CPU equipped with 48 GB of RAM over 15 random seeds, using existing Omnisafe implementations of the algorithms \cite{omnisafe}. For the environments (Fig. \ref{fig: envs_and_density}), we create two Unity-based scenarios considering a navigation task in a cluttered indoor environment with static obstacles for a Turtlebot3 mobile robot and a more challenging aquatic environmental monitoring task. The latter is characterized by an autonomous surface vehicle operating in a simulated aquatic environment with drift, sensor noise, and external disturbances. 

\subsection{Simulation Results}
 \begin{figure*}[t]

  \begin{minipage}[c]{0.6\textwidth}
    \centering
    \includegraphics[width=\linewidth]{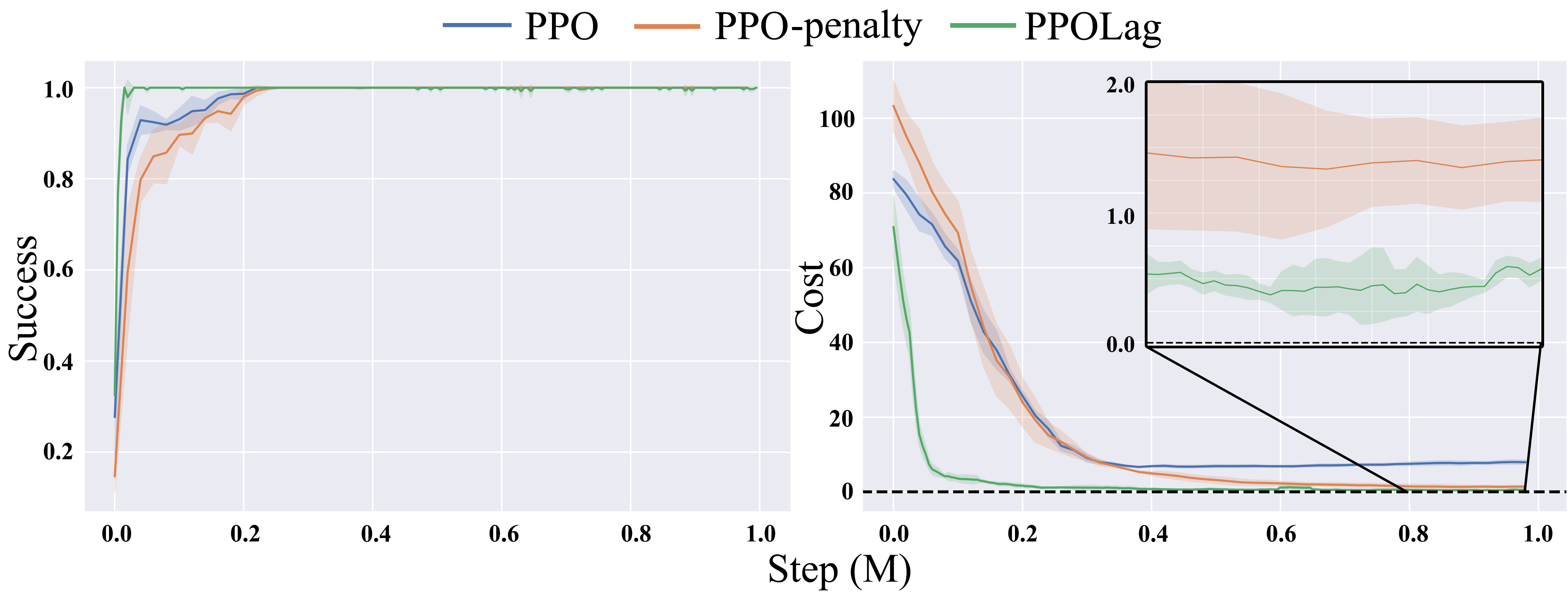}
  \end{minipage}%
  \hspace{1mm}
    \begin{minipage}[c]{0.35\textwidth}
    \centering
    \vspace{0.5em}
     \scriptsize
       



    \begin{tabular}{lcc}
    \multicolumn{2}{l}{\textbf{TB3 Evaluation}}&\\
    \toprule
    \textbf{Method} & \textbf{Success (\%)} & \textbf{Collision (\%)} \\
    \midrule
    PPO & 100$\pm$0 \% & 14.99 $\pm$0.02\% \\
    PPO+CBF & \textbf{100$\pm$0 \%} & \textbf{0$\pm$0\%} \\
    \midrule
    PPO\_penalty & 100$\pm$0 \% & 3.32 $\pm$1.03\% \\
    PPO\_penalty+CBF & \textbf{100$\pm$0 \%} & \textbf{0$\pm$0\%}   \\
    \midrule
    PPOLag & 99.66 $\pm$ 0.47\% & 0.47 $\pm$ 0.21\%\\
    PPOLag+CBF & \textbf{100$\pm$0 \%} & \textbf{0$\pm$0\%} \\
    \bottomrule
    \end{tabular}

  \end{minipage}

  \vspace{1em}

  \begin{minipage}[c]{0.6\textwidth}
    \centering
    \includegraphics[width=\linewidth]{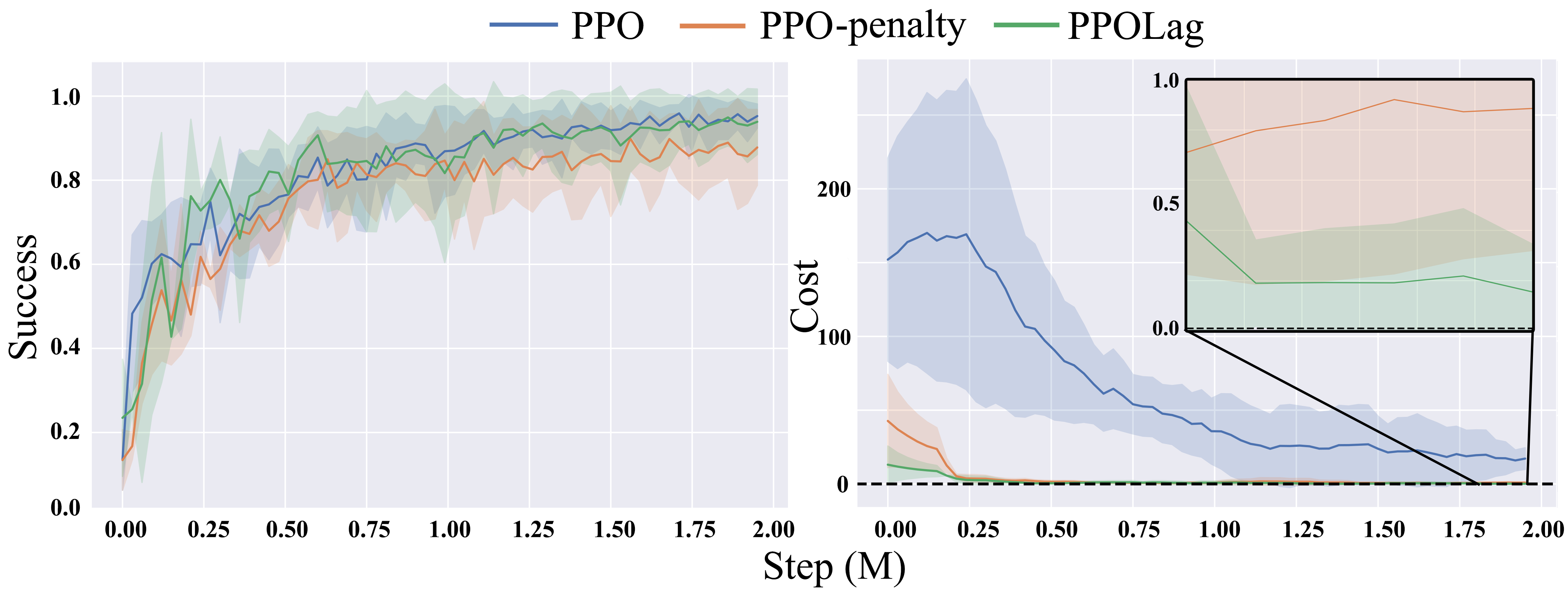}
  \end{minipage}%
  \hspace{2mm}
 \begin{minipage}[c]{0.35\textwidth}
    \centering
    \vspace{0.5em}
     \scriptsize
    \begin{tabular}{lcc}
     \multicolumn{2}{l}{\textbf{Aquatic Evaluation}}&\\
    \toprule
    \textbf{Method} & \textbf{Success (\%)} & \textbf{Collision (\%)} \\
    \midrule
    PPO & 84.2 $\pm$ 9.75 \% & 11.22 $\pm$ 3.26 \%\\
    PPO+CBF & \textbf{87.0 $\pm$ 12.72\%}& \textbf{0.0 $\pm$ 0.0} \%  \\
    \midrule
    PPO\_penalty & 89.7 $\pm$ 5.19 \% & 0.007 $\pm$ 0.005 \%\\
    PPO\_penalty+CBF & \textbf{96.7 $\pm$ 2.30\%}& \textbf{0.0 $\pm$ 0.0} \%  \\
    \midrule
    PPOLag & 84.0 $\pm$ 5.65 \% & 0.001 $\pm$ 0.002 \%\\
    PPOLag+CBF & \textbf{94.0 $\pm$ 4.24\%}& \textbf{0.0 $\pm$ 0.0} \% \\
    \bottomrule
    \end{tabular}
  \end{minipage}
 
  \caption{Training and evaluation results. On the left, we report the learning phase using PPO, PPO\_penalty, and PPOLag, respectively. On the right, we report the evaluation results of the proposed verification-guided CBF on the trained policies over 100 trajectories.}
  \label{fig: exp_results}

\end{figure*}
Our results are reported in Fig. \ref{fig: exp_results}. During training, all methods achieve a high success rate, indicating successful navigation behaviors towards random target locations. However, collisions still occur at convergence for all the DRL baselines. The average cost at convergence for the safety-oriented methods---PPOLag and PPO\_penalty approaches---is significantly closer to zero compared to the unconstrained baseline. Notably, even Lagrangian-based methods fail to fully satisfy the imposed strict cost threshold. The benefits of our CBF layer are clearly highlighted in the tables of Fig. \ref{fig: exp_results} (right) - all the given (unsafe) DRL policies combined with our method have a higher success rate while achieving safe navigation with zero collisions. This result confirms both the effectiveness and the potential of our framework in guaranteeing safety while mitigating instances where the agent gets stuck or exhibits near-zero cost during training but fails to be deployed safely. This phenomenon is evident in the aquatic navigation task, where both PPO\_penalty and PPOLag achieve low collision rates (at the expense of lower success rates) compared to the unconstrained PPO baseline. Crucially, our method demonstrates strong recovery capabilities in these situations, enabling the agent to progress toward goal locations. 

This behavior is further highlighted in Fig. \ref{fig: envs_and_density} (right), which shows enumerated regions where the agent not only tends to collide but also moves with near-zero linear velocity. Interestingly, some of these red-marked areas lack any visible obstacles, suggesting that the agent gets stuck even in open space due to suboptimal local policy decisions. It is important to note that our approach does not endow the agent with navigation skills; if the base policy lacks the ability to reach the goal from certain states, our method is not designed to compensate for this. Rather, our approach enhances the base policy by helping the agent recover from stuck situations, and reaching the goal is due only to the policy's capabilities to generalize to unseen scenarios properly. 

\subsection{Real-World Results}
\begin{figure}[h!]
    \centering
  \begin{minipage}[c]{0.45\textwidth}
    
    \includegraphics[width=0.8\linewidth]{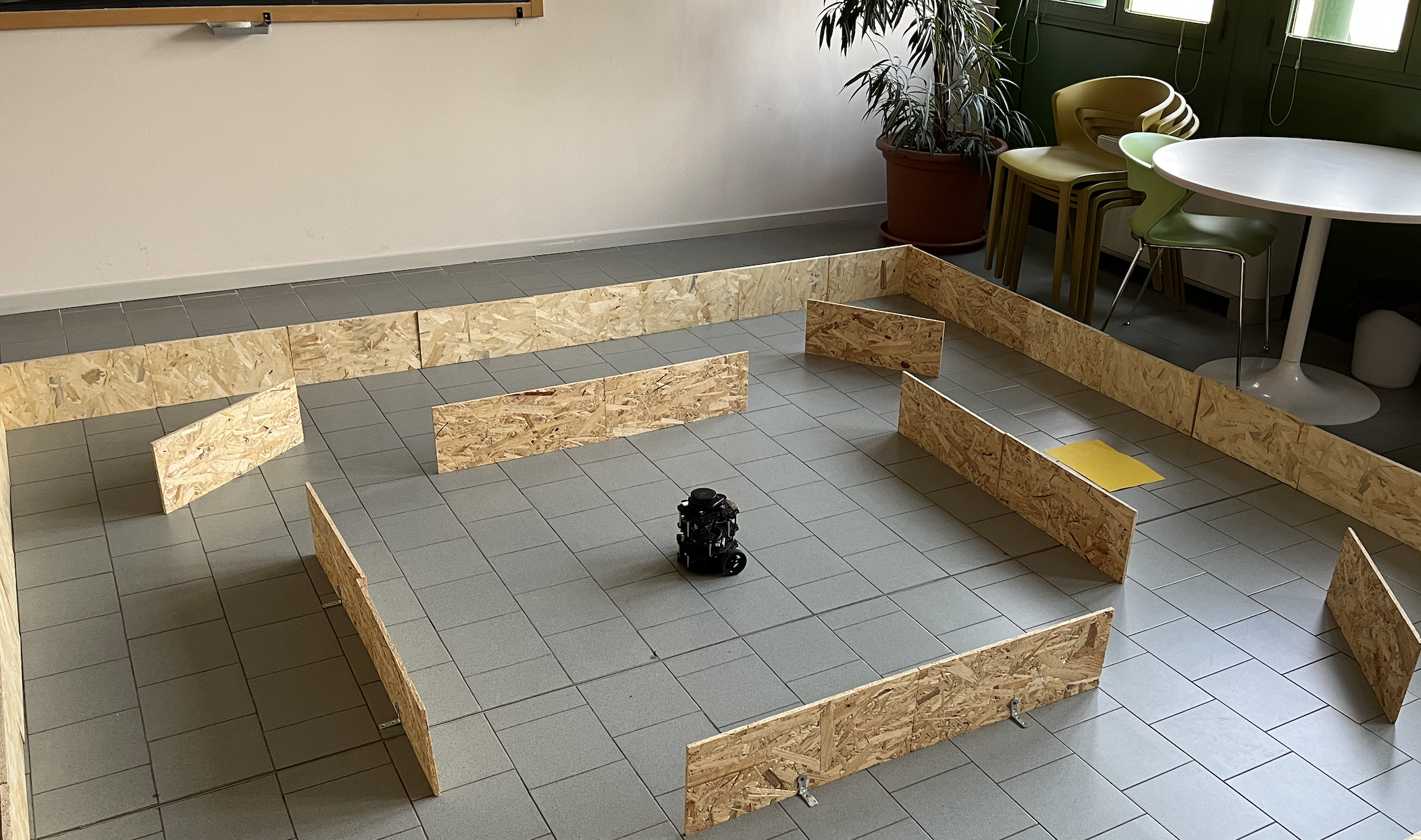} 
  \end{minipage}
  \hfill
  \begin{minipage}[c]{0.45\textwidth}
    \begin{tabular}{cc}
    \toprule
    \textbf{Success Rate (\%)} & \textbf{Collision Rate (\%)} \\
    \toprule
    72.7 $\pm$ 5.19 \% & 24.9 $\pm$ 2.1 \%\\
    \textbf{100 $\pm$ 0.0\%}& \textbf{0.0 $\pm$ 0.0} \%  \\
    \bottomrule
    \end{tabular}
       \end{minipage}
  \caption{Real-world experiments setup with PPOLag (first row) and PPOLag+CBF (second row).}
  \label{fig:real_exp}

\end{figure}

On top of the simulation results, Unity allows us to deploy the policies onto ROS2-enabled platforms, such as the Turtlebot3 robot. We thus selected the three best-performing seeds of the PPOLag model at convergence (i.e., the baseline achieving the best trade-off between safety and performance) and evaluated the effectiveness of our CBF-based approach in the real setting depicted on the left of Fig. \ref{fig:real_exp}. The right side of the figure reports the average results obtained from ten trajectories. These results confirm that our approach consistently achieves zero constraint violations while maintaining effective navigation performance, even under real-world sensor readings. The complete video of our empirical evaluation is available here \url{https://drive.google.com/file/d/1xMJQaRCegKjDR7FQvksD5Sxf2-JCtF1a/view?usp=share_link}.

\section{Conclusion}

In this work, we proposed a novel framework that integrates probabilistic verification to design a CBF-based layer that ensures safety for given DRL policies trained for autonomous mapless navigation. 
By identifying unsafe regions in the state space through probabilistic enumeration and enforcing safety constraints via CBF-based control, our approach provides a principled and scalable method to correct unsafe actions while preserving and potentially improving the performance of the learned policies. 
Unlike traditional methods that rely heavily on domain-specific knowledge or incur high computational costs, our framework is agnostic to the deployment environment and operates independently of the DRL training process, enhancing its general applicability.
Through extensive evaluations in both simulated environments and real-world deployment scenarios, we demonstrated that our approach achieves robust safety guarantees with zero violations of safety constraints. At the same time, it preserves effective navigation behaviors across different robot platforms, including an indoor robot ground and an aquatic drone. 
These results confirm the potential of combining probabilistic safety analysis and CBF-based control as a practical path forward for deploying safe, adaptive, and generalizable autonomous agents in complex and uncertain environments.
Future work will explore extending this framework to multi-agent settings, dynamic environments, and broader classes of safety specifications, further pushing the boundary of safe and reliable reinforcement learning for real-world robotics deployment.


\bibliographystyle{unsrt}  
\bibliography{references}

\end{document}